\documentclass[runningheads]{llncs}
\usepackage{graphicx}
\usepackage{comment}
\usepackage{amsmath,amssymb} 
\usepackage{color}
\usepackage{booktabs}
\usepackage{subcaption}
\usepackage{lipsum}

\usepackage[width=122mm,left=12mm,paperwidth=146mm,height=193mm,top=12mm,paperheight=217mm]{geometry}

\begin{document}
\pagestyle{headings}
\mainmatter
\def\ECCVSubNumber{4574}  

\title{DCANet: Learning Connected Attentions for Convolutional Neural Networks } 


\titlerunning{DCANet}
%
\author{Xu Ma \and Jingda Guo \and Sihai Tang \and Zhinan Qiao \and Qi Chen \and Qing Yang \and Song Fu}
\authorrunning{Xu Ma.}
%
\institute{University of North Texas.  Denton, US}
\maketitle

\begin{abstract}
While self-attention mechanism has shown promising results for many vision tasks, it only considers the current features at a time. We show that such a manner cannot  take full advantage of the attention mechanism. In this paper, we present Deep Connected Attention Network (DCANet),  a novel design that boosts attention modules in a CNN model without any modification of the internal structure. To achieve this, we interconnect adjacent attention blocks, making  information flow among attention blocks possible. With DCANet, all attention blocks in a CNN model are trained jointly, which improves the ability of attention learning.
Our DCANet is generic. It is not limited to a specific attention module or base network architecture. Experimental results on ImageNet and MS COCO benchmarks show that DCANet consistently outperforms the state-of-the-art attention modules with a minimal additional computational overhead in all test cases. 
All code and models are made publicly available.

\end{abstract}

\section{Introduction}
In the last few years, we have witnessed a flourish of self-attention mechanism in the vision community.
As a common practice in self-attention design, the attention modules are integrated sequentially with each block in a base CNN architecture, in pursuit of an easy and efficient implementation. 
Benefiting from the inherent philosophy and this simple design, self-attention mechanism performs well in a diverse range of visual tasks~\cite{wang2017residual,anderson2018bottom,fu2019dual}.

In spite of the improvement achieved by the existing designs, a question we ask is: do we take full advantage of self-attention mechanism?  We can address this question from two aspects: human visual attention system and empirical insights from SENet~\cite{hu2018squeeze}.

Previous studies in the literature provided deep insights into the human visual attention system. In~\cite{ungerleider2000mechanisms}, experimental results indicate that two stimuli present at the same time in the human cortex are not processed independently. Instead they interact with each other. Moreover,  research in physiology discovers that human visual representations in the cortex are activated in a parallel fashion, and the cells participating in these representations are engaged by interacting with each other~\cite{chelazzi1998responses}.
These works show the {\em important interaction among attention units}. However, this critical property of human visual attention has not been considered in the existing designs of self-attention modules. Existing attention networks only include an attention block following a convolutional block, which makes the attention block only learn from current feature maps without sharing information with others. As a result, the independent attention blocks cannot effectively decide what to pay attention to.

\begin{figure}[!t]
\centering
\includegraphics[width=1.0\linewidth]{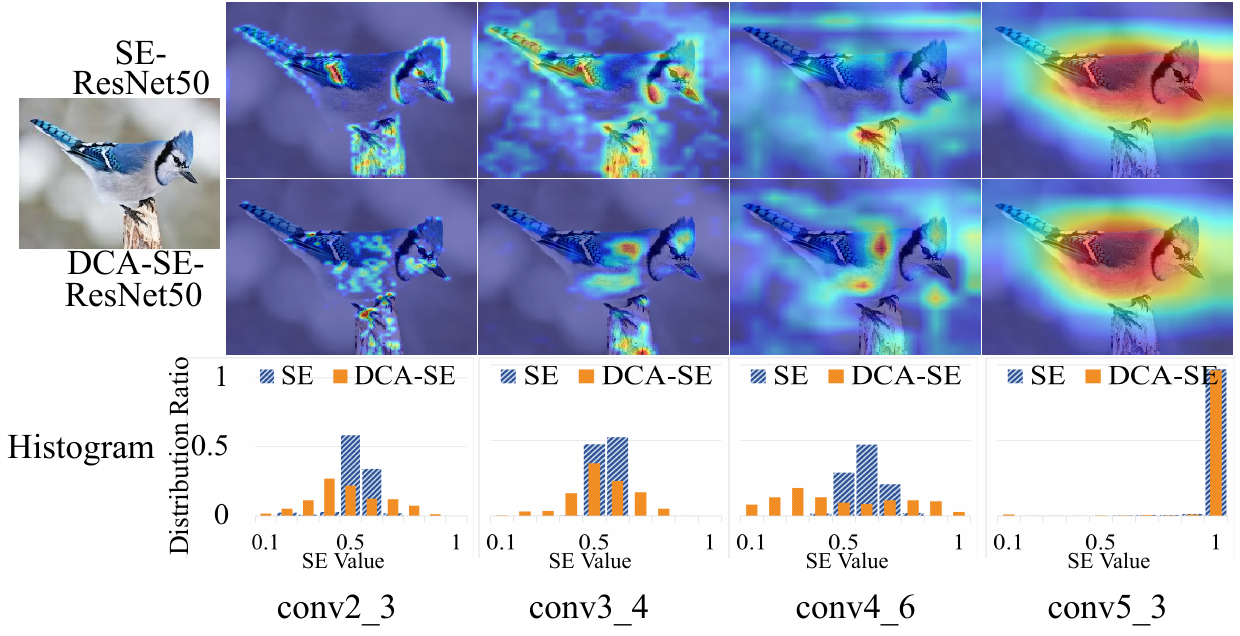}
\caption{Illustration of our DCANet. Top and middle lines: we visualize intermediate feature activation using Grad-CAM~\cite{selvaraju2017grad}. Vanilla SE-ResNet50 varies its focus dramatically at different stages. In contrast, our DCA enhanced SE-ResNet50 \textbf{progressively and recursively} adjusts focus, and closely pays attention to the target object. Bottom line: Corresponding histogram of SE attention values. Clearly, the values of SE are concentrated around 0.5, resulting in very little discrimination. With DCANet, the distribution becomes relatively uniform.}
\label{fig:1}
\end{figure}

Additionally, we study self-attention using SENet~\cite{hu2018squeeze} which is a simple attention network that investigates the channel relationships. We visualize the intermediate attention maps as shown in Fig.~\ref{fig:1} (top line) at each stage in SE-ResNet50~\cite{he2016deep}. Interestingly, we observe that SE block can hardly adjust the attention to the key regions, and it even changes focus dramatically at different stages. We plot a histogram of SE's attention value for each block, as shown in Fig.~\ref{fig:1} (bottom line). We find that SE's values cluster around 0.5, showing an insufficient learning ability of the attention modules. 
A reasonable explanation is that a lack of extra information in learning from self-attention affects its discrimination ability.
This in turn motivates us to connect attention blocks.

Both human visual attention and our study of SENet show an insufficient exploitation of the self-attention mechanism, and  a new design that allows attention blocks to cooperate with each other is desirable. In this paper, we present a Deep Connected Attention network (DCANet), aiming to address the problem. DCANet gathers information from precedent attention and transmits it to the next attention block, making attention blocks cooperate with each other, which improves attention's learning ability. Without any modification of the internal structure, DCANet introduces a chain of connections among attention blocks. It  can be applied to various self-attention modules, e.g., SENet~\cite{hu2018squeeze}, CBAM~\cite{woo2018cbam}, SKNet~\cite{li2019selective}, and more, regardless of the choice of base architecture. In Fig.~\ref{fig:1}, we demonstrate the ability of DCANet based on SE-ResNet50. We note that ResNet50 contains shortcut connections between adjacent blocks. However, those shortcut connections do not improve  attention learning. In contrast, DCANet incorporates attention connections among attention blocks, which is  different from shortcut connections between convolutional blocks.

DCANet is conceptually simple and generic and empirically powerful. We apply DCANet to multiple state-of-the-art attention modules and a number of base CNN architectures to evaluate its performance for visual tasks.  Without bells and whistles,  the DCA-enhanced networks \textbf{outperform all of the original counterparts}. For ImageNet 2012 classification~\cite{russakovsky2015imagenet},  DCA-SE-MobileNetV2 outperforms SE-MobileNetV2 by 1.19\%, with  negligible parameters and FLOPs increase.  We also employ the DCA-enhanced attention network as a backbone for object detection on the MS COCO dataset~\cite{lin2014microsoft}. Experimental results show that the DCANet-enhanced attention networks outperform the vanilla networks with different detectors.

\section{Related Work}
\textbf{Self-attention mechanisms.} Self-attention mechanism explores the interdependence within the input features for a better representation. 
In these years, attention mechanisms hold prevalence across a large range of tasks, from machine translation~\cite{bahdanau2014neural} in natural language processing to object detection~\cite{cao2019gcnet} in computer vision.
To the best of our knowledge, applying self-attention to explore global dependencies was first proposed in~\cite{vaswani2017attention} for machine translation.
More recently, self-attention has gathered much more momentum in the field of computer vision.
To investigate channel interdependencies, SENet~\cite{hu2018squeeze}, GENet~\cite{hu2018gather} and SGENet~\cite{SGENet} leverage self-attention for contextual modeling. For global context information, NLNet~\cite{wang2018non} and GCNet~\cite{cao2019gcnet} introduce self-attention to capture long-range dependencies in non-local operations. BAM~\cite{park2018bam} and CBAM~\cite{woo2018cbam} consider both channel-wise and spatial attentions.  Beyond channel and spatial dependencies, SKNet~\cite{li2019selective} applies self-attention to kernel size selection.

\vspace{0.1cm}
\noindent\textbf{Residual connections.}
The idea of Residual connection comes from~\cite{s1998accelerated}. By introducing a shortcut connection, neural networks are decomposed into biased and centered subnets to accelerate gradient descent. ResNet~\cite{he2016deep,he2016identity} adds an identity mapping to connect the input and output of each convolutional block, which drastically alleviates the degradation problem~\cite{he2016deep} and opens up the possibility for deep convolutional neural networks. 
Instead of connecting adjacent convolutional blocks, DenseNet~\cite{huang2017densely} connects each block to every other block in a feed-forward fashion. 
FishNet~\cite{sun2018fishnet} connects layers in pursuit of propagating gradient from deep layers to shallow layers. 
DLA~\cite{yu2018deep} shows that residual connection is a common approach of layer aggregation.
By iteratively and hierarchically aggregating layers in a network, DLA is able to reuse feature maps generated by each layer.

Despite the fact that residual connections have been well studied for base network architectures, they are still fairly new when it comes to integration with attention mechanisms.
For example, RANet~\cite{wang2017residual} utilizes residual connections in attention block; in~\cite{wang2017residual,park2018bam}, residual learning is used in attention modules to facilitate the gradient flow. In  contrast to leveraging residual connection \textit{in} attention blocks, we explore residual connections \textit{between} attention blocks. 

\vspace{0.1cm}
\noindent\textbf{Connected Attention.}
Recently, there has been a growing interest for building connections in attention blocks.
%
In~\cite{fu2017look}, a new network structure named RA-CNN is proposed for fine-grained image recognition; RA-CNN recurrently generates attention region based on current prediction to learn the most discriminative region. 
By doing so, RA-CNN obtains an attention region from coarse to fine. 
In GANet~\cite{arXiv2019GANet}, the top attention maps generated by customized background attention blocks are up-sampled and sent to bottom background attention blocks to guide attention learning.
Different from the recurrent and feed-backward methods, our DCA module enhances attention blocks in a feed-forward fashion, which is more computation-friendly and easier to implement.

\section{Deep Connected Attention}
\begin{figure*}[!t]
\centering
\includegraphics[width=1\linewidth]{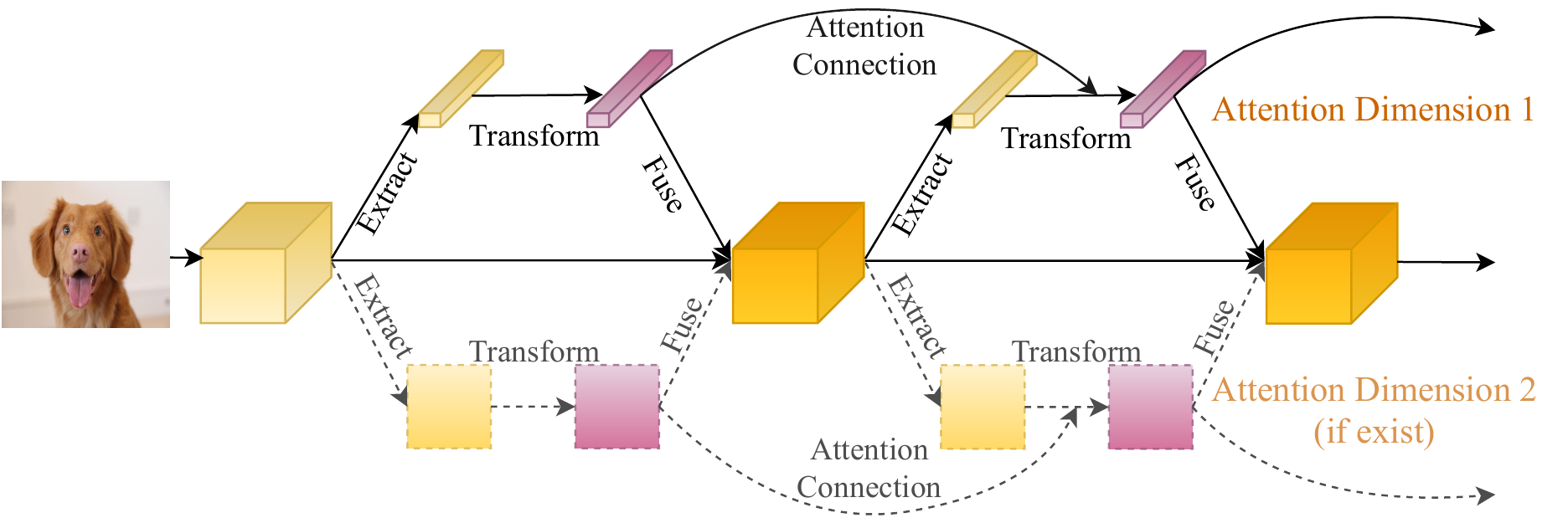}
\caption{\textbf{An overview of our Deep Connected Attention Network.}
We connect the output of transformation module in the previous attention block to the output of extraction module in current attention block.
In the context of multiple attention dimensions, we connect attentions along each dimension. Here we show an example with two attention dimensions. It can be one, two or even more attention dimensions.
}
\label{fig:framework}
\end{figure*}

Deep Connected Attention Network is conceptually simple but empirically powerful. 
By analyzing the inner structure of various attention blocks, we propose a generic connection scheme that not confined to particular attention blocks.
We merge the previous attention features and current extracted features by parameterized addition to ensure the information flow among all attention blocks in a feed-forward manner and prevent attention information from varying a lot in each step. 
Fig.~\ref{fig:framework} illustrates the overall pipeline of our method. 

\subsection{Revisiting Self-Attention Blocks}
We first revisit several prevalent attention modules to analyze the inner structure.
As a common practice, we boost the base CNN architecture by adding extra attention blocks laterally. However, different attention blocks are tailored for different purposes, the implementations are also diverse. For instance, SE block composes of two fully-connected layers, while GC block includes several convolutional layers. Therefore, it is not easy to directly provide a standard connection schema that is generic enough to cover most attention blocks. To tackle this problem, we study state-of-the-art attention blocks and summarize their processing and components. 

\begin{figure}[!t]
\centering
\includegraphics[width=0.9\linewidth]{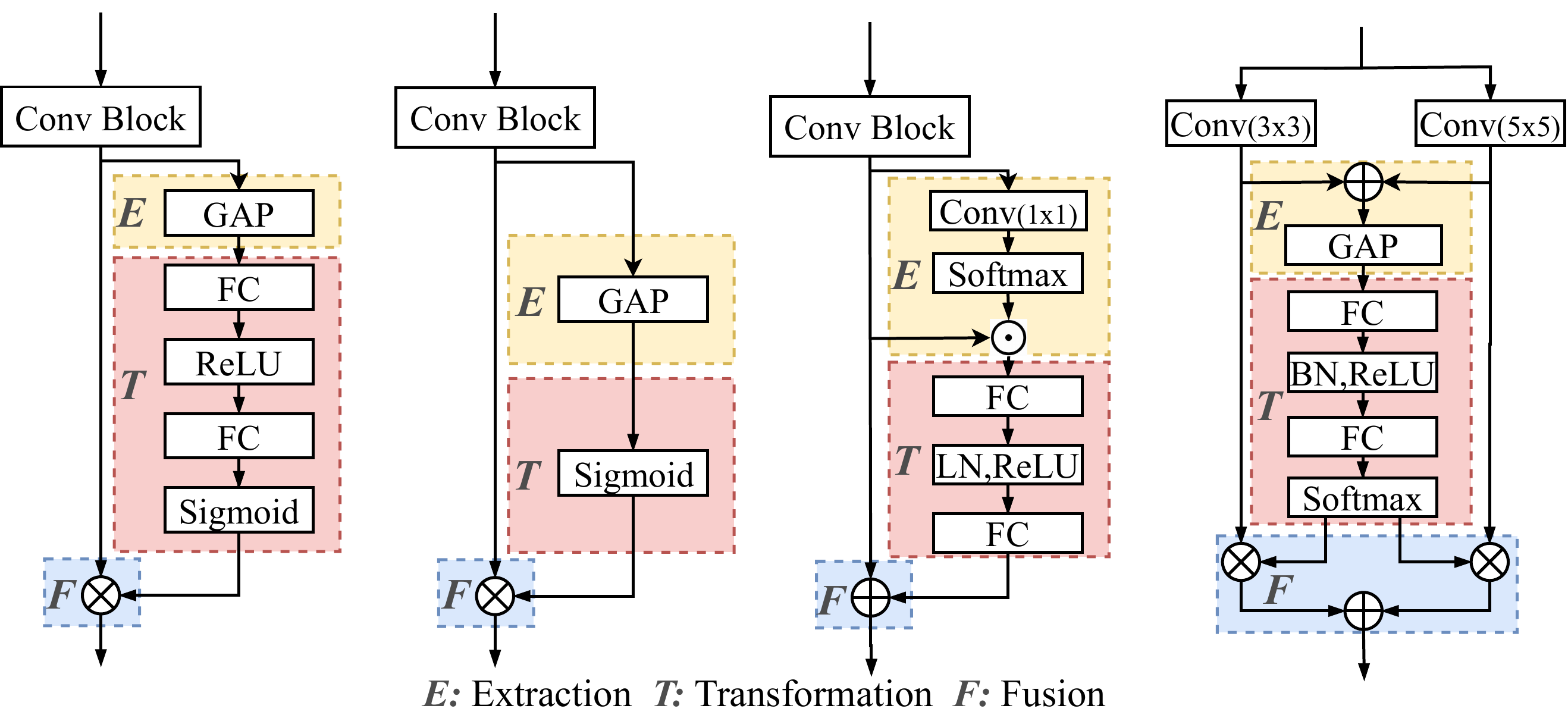}
\caption{We model an attention block by three components: Feature extraction, Transformation and Fusion.  \textbf{ From left to right:} SE block~\cite{hu2018squeeze}, GE$\theta^-$ block~\cite{hu2018gather}, GC block~\cite{cao2019gcnet}, and SK block~\cite{li2019selective}. ``$\oplus$" denotes element-wise summation, ``$\otimes$" represents element-wise multiplication, and ``$\odot$" performs matrix multiplication. } 
\label{fig:example}
\end{figure}

Inspired by recent work~\cite{Empirical_Attention_Summary,cao2019gcnet,LCTNet} that formulate attention modules and their components (focus on SENet and NLNet), we study various attention modules and develop a generalized attention framework, in which
an attention block consists of three components: context extraction, transformation, and fusion.
Extraction serves as a simple feature extractor, transformation transforms the extracted features to a new non-linear attention space, while fusion merges attention and original features. 
These components are generic and not confined to a particular attention block. Figure~\ref{fig:example} exemplifies four well-known attention blocks and their modeling by using the three components.  

\textbf{Extraction} is designed for gathering feature information from a feature map.  For a given feature map $\mathbf{X} \in \mathbb{R}^{C\times W\times H}$ produced by a convolutional block, we extract features from $\mathbf{X}$ by an extractor $g$: $\mathbf{G} =g \left(\mathbf{X},w_g\right)$,  where $w_g$ is parameter for the extraction operation and $\mathbf{G}$ is the output. When $g$ is a parameter-free operation, $w_g$ is not needed (like pooling operations).
The flexibility of $g$ makes $\mathbf{G}$ take different shapes depending on the extraction operation. For instance, SENet and GCNet gather feature map $\mathbf{X}$ as a vector ($\mathbf{G}\in \mathbb{R}^C$)
while the spatial attention module in CBAM gathers feature map to a tensor ($\mathbf{G}\in \mathbb{R}^{2\times W \times H}$). 

\textbf{Transformation} processes the gathered features from extraction and transforms them into a non-linear attention space. Formally, we define $t$ as a feature transformation operation, and the output of an attention block can be expressed as $\mathbf{T} = t \left(\mathbf{G},w_t\right)$. Here ${w_t}$ denotes parameters used in the transform operation, and $\mathbf{T}$ is the output of the extraction module. 

\textbf{Fusion} integrates the attention map with the output of the original convolutional block.  An attention guided output $\mathbf{X}'\in \mathbb{R}^{C \times W\times H}$ can be presented as $
{\mathbf{X}'}_i  = \mathbf{T}_i \circledast \mathbf{X}_i
$,
where $i$ is the index in a feature map and ``$\circledast$" denotes a fusion function; ``$\circledast$" performs element-wise multiplication when the design is scaled dot-product attention~\cite{li2019selective,hu2018squeeze,woo2018cbam}, and summation otherwise~\cite{cao2019gcnet}.

\subsection{Attention Connection}
\label{section: attention_connection}
Next, we present a generalized attention connection schema by using the preceding attention components. Regardless of the implementation details, an attention block can be modeled as:
\begin{equation}
\mathbf{X}' =  t\left(g\left(\mathbf{X},w_g\right),w_t\right)
\circledast
\mathbf{X}.
\end{equation}

%
As explained in the previous section, the attention maps generated by the transformation component is crucial for attention learning.
To construct connected attention, we feed the previous attention map to the current transformation component, which merges previous transformation output and the current extraction output.
\textbf{This connection design ensures the current transformation module learns from both extracted features and previous attention information.}
The resulting attention block can be described as:
\begin{equation}
\mathbf{X}' = t\left(f\left(\alpha\mathbf{G},\beta\tilde{\mathbf{T}}\right),w_t\right)
\circledast
\mathbf{X},
\label{equation: connection}
\end{equation}
where $f\left(\cdot \right)$ denotes the connection function, $\alpha$ and $\beta$ are learnable parameters, and  $\tilde{\mathbf{T}}$ is the attention map generated by the previous attention block. In some cases (e.g., SE block and GE block), $\tilde{\mathbf{T}}$ is scaled to the range of $\left(0,1\right)$. For those attention blocks, we multiply $\tilde{\mathbf{T}}$
by $\tilde{\mathbf{E}}$ to match the scale, where $\tilde{\mathbf{E}}$ is the output of the Extraction component in the previous attention block. We also note that if $\alpha$ is set to 1 and $\beta$ is set to 0, the attention connections are not used and the DCA enhanced attention block is reduced to the vanilla attention block. That is the vanilla network is a special case of our DCA enhanced attention network.

Next, we present two schemas that instantiate the connection function $f\left(\cdot\right)$.


\vspace{1mm}
\noindent \textbf{Direct Connection.}
We instantiate the $f\left(\cdot\right)$ by adding the two terms directly.
The connection function can be presented as:

{\small
\begin{equation}
\label{equ:direct}
f\left(\alpha\mathbf{G}_i,\beta\tilde{\mathbf{T}}_i\right)
= \alpha\mathbf{G}_i + \beta\tilde{\mathbf{T}}_i,
\end{equation}
}

\noindent where $i$ is the index of a feature. In Equation (\ref{equ:direct}), $\tilde{\mathbf{T}}$ can be considered as an enhancement of $\mathbf{G}$.

\vspace{1mm}
\noindent\textbf{Weighted Connection.} Direct connection can be augmented by using weighted summation. To avoid introducing extra parameters, we calculate weights using $\alpha\mathbf{G}$ and $\beta\tilde{\mathbf{T}}$.  The connection function is represented as
{\small
\begin{equation}
f\left(\alpha\mathbf{G}_i,\beta\tilde{\mathbf{T}}_i\right)
= \frac{\left |  \alpha\mathbf{G}_i\right |^2 }{\alpha\mathbf{G}_i+\beta\tilde{\mathbf{T}}_i}
+ 
\frac{\left | \beta\tilde{\mathbf{T}}_i \right |^2}{\alpha\mathbf{G}_i+\beta\tilde{\mathbf{T}}_i}.
\end{equation}
}
Compared to the direct connection, the weighted connection introduces a competition between $\alpha\mathbf{G}$ and $\beta\tilde{\mathbf{T}}$. Besides, it can be easily extended to a softmax form, which is more robust and less sensitive to trivial features.

Experimental results from ablation studies (presented in Table \ref{tab:ablation_connection}) show that the result is insensitive to the connection schemas, indicating that the performance improvement comes more from connections between attention blocks than the specific form of the connection function.
Thus, we use a direct connection in our method by default. 

\subsection{Size Matching}
\label{section:size_matching}
Feature maps produced at different stages in a CNN model may have different sizes. Thus, the size of the corresponding attention maps may vary as well, and such a mismatch makes our DCANet impossible to be applied between the two stages.
In order to tackle this issue, we match the shape of attention maps along the channel and spatial dimensions adaptively.

For the channel, we match sizes using a fully-connected layer (followed by layer normalization \cite{ba2016layer} and ReLU activation) to convert $C'$ channels to $C$ channels, where $C'$ and $C$ refer to the number of previous and current channels, respectively.
Omitting biases for clarity, parameters introduced for channel size matching is $C'\times C$. 
To further reduce parameter burdens in attention connections, we re-formulate the direct fully-connected layer by two lightweight fully-connected layers; the output sizes are $C/r$ and $C$, respectively, where $r$ is reduction ratio. This modification significantly reduces the number of introduced parameters. The influences of channel size matching strategies can be found in Table~\ref{tab:ablation_size_matching_channel}. In all our experiments, we use two fully-connected layers with $r=16$ to match channel size, unless otherwise stated. 

To match the spatial resolutions, a simple yet effective strategy is to adopt an average-pooling layer. We set stride and receptive field size to the scale of resolution reduction. Max-pooling also works well in our method, but it only considers parts of the information instead of the whole attention information. In addition to pooling operations, an alternative solution is the learnable convolutional operation.
However, we argue that it is not suitable for our purpose as it introduces many parameters and does not generalize well. 
Detail ablation experiments on spatial resolution size matching can be found in Table~\ref{tab:ablation_size_matching_spatial}.

\subsection{Multi-dimensional attention connection}
\label{section:multi_dimension}
We note that some attention blocks focus on more than one attention dimension. For instance, BAM~\cite{park2018bam} and CBAM~\cite{woo2018cbam} infer attention maps along both channel and spatial dimensions. Inspired by  Xception~\cite{chollet2017xception} and MobileNet~\cite{howard2017mobilenets,sandler2018mobilenetv2}, we design attention connections for one attention dimension at a time. To build a multi-dimensional attention block, we connect attention maps along with each dimension and assure connections in different dimensions are independent of one another (as shown in Fig.~\ref{fig:framework}). This decoupling of attention connections brings two advantages: 1) it reduces the number of parameters and computational overhead; 
2) each dimension can focus on its intrinsic property. 

\section{Experiments}

In this section, we exhaustively evaluate DCANet for image recognition and object detection tasks. Experimental results on ImageNet~\cite{russakovsky2015imagenet} and MS-COCO~\cite{lin2014microsoft} benchmarks demonstrate the effectiveness of our method. Comprehensive ablation studies are presented to thoroughly investigate the internal properties.


\subsection{Classification on ImageNet}

\begin{table*}[!t]
\caption{Single-crop classification accuracy (\%) on ImageNet validation set. We re-train all models and report results in the ``re-implement" column. The corresponding DCANet variants are presented in the ``DCANet" column.  The best performances are marked as \textbf{bold}. ``-" means no experiments since our DCA is designed for attention blocks, which are not existent in base networks.}
\begin{center}
\begin{tabular*}{1\textwidth}{c @{\extracolsep{\fill}}lcccccccc}
\toprule
& \multicolumn{4}{c}{Re-implementation} & \multicolumn{4}{c}{DCANet}\\\cline{2-9}
& Top-1 & Top-5 & GFLOPs & Params & Top-1 & Top-5 & GFLOPs & Params \\
\midrule
MoibleNetV2~\cite{sandler2018mobilenetv2} &71.03   &90.07 &0.32  &3.50M  &- &-  &- &- \\
+ SE~\cite{hu2018squeeze} &72.05   &90.58  &0.32  &3.56M  &\textbf{73.24} &91.14  &0.32 &3.65M\\
+ SK~\cite{li2019selective} &74.05   &91.85  &0.35   &5.28M  &\textbf{74.45} & 91.85 & 0.36 &5.91M\\
+ GE$\theta^-$~\cite{hu2018gather} &72.28&90.91&0.32&3.50M &  \textbf{72.47} &90.68 &0.32 &3.59M\\

+ CBAM~\cite{woo2018cbam} &71.91   &90.51  & 0.32  &3.57M  &\textbf{73.04} &91.18  &0.34 &3.65M\\ 
\midrule
Mnas1\_0~\cite{tan2019mnasnet} &71.72   &90.32  &0.33  &4.38M  &- & - &- &- \\
+ SE~\cite{hu2018squeeze} &69.69   &89.12  & 0.33 &4.42M  &\textbf{71.76} &90.40  &0.33 &4.48M \\
+ GE$\theta^-$~\cite{hu2018gather} &72.72&90.87&0.33&4.38M  & \textbf{72.82} & 91.18&0.33 &4.48M\\

+ CBAM~\cite{woo2018cbam} &69.13   &88.92  &0.33   &4.42M  &  \textbf{71.00}& 89.78 &0.33 & 4.56M \\
\midrule

ResNet50~\cite{he2016deep} &75.90   &92.72  &4.12  &25.56M  &- &-  &-&- \\
+ SE~\cite{hu2018squeeze} &77.29   &93.65  &4.13  &28.09M  &\textbf{77.55} &93.77  &4.13 &28.65M \\
+ SK~\cite{li2019selective} &77.79   &93.76  &5.98   &37.12M  &\textbf{77.94} &93.90  &5.98 &37.48M \\
+ GE$\theta^-$~\cite{hu2018gather} &76.24   &92.98  &4.13   &25.56M  &\textbf{76.75}  &93.36  &4.13 &26.12M\\
+ CBAM~\cite{woo2018cbam} &77.28   &93.60  &4.14   &28.09M  &\textbf{77.83} &93.72  &4.14 & 30.90M\\

\bottomrule

\end{tabular*}
\end{center}
\label{table:comparison_ImageNet}
\end{table*}

We apply our DCANet to a number of state-of-the-art attention blocks, including SE~\cite{hu2018squeeze}, SK~\cite{li2019selective}, GE~\cite{hu2018gather}, and CBAM~\cite{woo2018cbam}. We use ResNet50~\cite{he2016deep} as the base network for illustration. 
As lightweight CNN models attract increasing attention due to their 
efficiency
on mobile devices, we also experiment on lightweight models, \textit{e.g.} MobileNetV2~\cite{sandler2018mobilenetv2}, to evaluate the performance of DCANet.
Additionally, we select MnasNet1\_0~\cite{tan2019mnasnet} as an example method on neural architecture search. We integrate DCA module with the original attention networks and measure the performance improvement on image classification. 

We train all models on the ImageNet 2012 training set and measure the single-crop ($224\times224$ pixels) top-1 and top-5 accuracy on the validation set. Our implementations are based on PyTorch~\cite{paszke2017automatic}. For training ResNet and variants, we use the setup in~\cite{he2016deep}.  We train models for 100 epochs on 8 Tesla V100 GPUs with 32 images per GPU (the batch size is 256). All models are trained using synchronous SGD with Nesterov momentum~\cite{sutskever2013importance} of 0.9 and a weight decay of 0.0001. The learning rate is set to 0.1 initially and lowered by a factor of 10 every 30 epochs. For lightweight models like MnasNet and MobileNetV2, we take cosine decay method~\cite{loshchilov2016sgdr} to adjust the learning rate and train the models for 150 epochs with 64 images per GPU.

Table~\ref{table:comparison_ImageNet} presents the detailed results on the validation set.
We empirically observed that integrating the DCA module improves the classification accuracy in all cases when compared to the vanilla attention models. 
Of note is that we are comparing with corresponding attention networks, which is stronger than the base networks. 
Among the tested networks, DCA-CBAM-ResNet50 improves the top-1 accuracy by 0.51\% compared with CBAM-ResNet50, and DCA-SE-MobileNetV2 improves the top-1 accuracy by 1.19\% compared with SE-MobileNetV2, but the computation overhead is comparable. The improvement demonstrates the efficiency of our DCANet.  

It surprises us to see that directly applying SE and CBAM attention blocks to MnasNet decreases the performance. One explanation is that MnasNet1\_0 (from PyTorch~\cite{paszke2017automatic}) is a model obtained from pre-defined network search space while SE and CBAM attention blocks (which contain learnable layers, GE$\theta^-$ contains no learnable layers) are not in the search space of MnasNet. However, when applying our DCANet, MnasNet with attention blocks can achieve comparable performance on par with the original MnasNet1\_0. 
This is because our DCA module propagates attention maps along multiple attention blocks and combines them together to achieve better classification.
Note in Equation (\ref{equation: connection}),  $\alpha$ and $\beta$ are learnable parameters, which dynamically integrates the influence of previous attention block with that of the current one, even suppress later attention models. For example, when $\beta$ is significant, $f\left(\cdot\right)$ biases toward the previous attention as the previous attention dominates.

\subsection{Ablation Evaluation}
In this subsection, we present and discuss experimental results from ablation tests on the ImageNet dataset for a deeper understanding of our DCANet.

\begin{table}[!t]
\caption{Ablation studies on ImageNet 2012 validation set.}\label{tbl:main}

\scriptsize
\centering
\begin{subtable}[t]{0.45\linewidth}
\caption{DCA connection schemas.}
\label{tab:ablation_connection}
\centering
\vspace{0pt}
\begin{tabular}{lcccc}
\toprule
Model & Top-1 & Top-5 & GFLOPs & Params \\
\midrule
SE &77.29 &93.65 &4.13 &28.09M\\ 
\midrule
+Direct &\textbf{77.55} &\textbf{93.77}  &4.13 &28.65M \\
+Softmax &77.52 &93.71 &4.13 &28.65M\\
+Weighted&77.49 &93.69 &4.13 &28.65M\\ 
\bottomrule
\end{tabular}

\caption{Multiple Attention dimensions.}
\label{tab:ablation_multi-dimension}
\begin{tabular}{lcccc}
\toprule
Model & Top-1 & Top-5 & GFLOPs & Params \\
\midrule
CBAM &77.28   &93.60  &4.14   &28.09M \\ 
\midrule
+DCA-C &77.79 &93.71 &4.14 &30.90M \\ 
+DCA-S &77.58 &\textbf{93.80} &4.14 &28.09M \\
+DCA-All &\textbf{77.83} & 93.72  &4.14 &30.90M \\
\bottomrule
\end{tabular}

\caption{Spatial size matching.}
\label{tab:ablation_size_matching_spatial}
\begin{tabular}{lcccc}
\toprule
Model & Top-1 & Top-5 & GFLOPs & Params \\
\midrule
CBAM &77.28 &93.60 &4.14 &28.09M \\ \midrule
Max Pooling &77.43 &93.77 &4.14 &28.09M \\ 
Avg Pooling &\textbf{77.58} &\textbf{93.80} &4.14 &28.09M \\ \bottomrule
\end{tabular}

\end{subtable}\hfill
\begin{subtable}[t]{0.5\linewidth}
\centering
\vspace{0pt}
\caption{Comparison of different depth. * indicates results from AANet~\cite{bello2019attention}.}
\label{tab:ablation_depth}
\begin{tabular}{lcccc}
\toprule
Model & Top-1 & Top-5 & GFLOPs & Params \\ \midrule
ResNet50 &75.90 &92.72 &4.12 &25.56M \\
SE &77.29 &93.65 &4.13 &28.09M  \\
DCA-SE &\textbf{77.55} &93.77 &4.13 &28.65M \\ \midrule

ResNet101 &77.87 &93.80 &7.85 &44.55M \\
SE &78.34 &94.16  &7.86  &49.33M  \\
DCA-SE &\textbf{78.45} &94.27 &7.86 &49.93M \\ \midrule

ResNet152* &78.4 &94.2 &11.58 &60.19M \\
SE &78.52 &94.07 &11.60 &66.82M  \\
DCA-SE &\textbf{78.61} &94.24 &11.60 &67.45M \\ \bottomrule
\end{tabular}

\caption{Channel matching based on SE-ResNet50. ``$r$" is reduction rate.}
\label{tab:ablation_size_matching_channel}
\begin{tabular}{lcccc}
\toprule
Model & Top-1 & Top-5 & GFLOPs & Params \\
\midrule
SE &77.29 &93.66 &4.13 &28.09M \\ \midrule
1 FC &\textbf{77.64} &93.74 &4.13 &30.90M \\ 
2 FC (r=16) &77.55 &\textbf{93.77} &4.13 &\textbf{28.65M} \\ 
2 FC (r=8) &77.50 &93.72 &4.13 &29.87M \\ 
2 FC (r=4)&77.42 &93.75 &4.13 &32.31M \\ \bottomrule 
\end{tabular}

\end{subtable}

\end{table}

\vspace{0.1cm}
\noindent \textbf{Connection Schema.}
As shown in Table~\ref{tab:ablation_connection},
all three connection schemas outperform vanilla SE-ResNet50. This indicates the performance improvement comes from the connections between attention blocks, rather than particular connection schema.
Besides, we observed little or nothing different in the top-1 and top-5 accuracy of these three connection schemas (77.55\% \textit{vs.} 77.52\% \textit{vs.} 77.49\%). By default we use direct connection which eases the implementation compared to others.

\vspace{0.1cm}
\noindent \textbf{Network Depth.}
We first compare SE-ResNet50 against the DCA boosted counterpart and then increase the depth from 50 to 101 and 152. 
Table~\ref{tab:ablation_depth} lists the experimental results. In the table, we can see that among the three networks, DCA-SE-ResNet always outperforms SE-ResNet and ResNet. For instance, our DCA-SE-ResNet outperforms ResNet by 1.65\%, 0.58\%, and 0.21\% when the network depth is 50, 101, and 152 respectively. However, the performance gain becomes smaller as the depth increases. 
A similar trend also occurs when applying SE module to ResNet. The key insight behind these phenomena is that applying attention modules to deep networks results in less performance improvement than applying to shallow networks, since the performance of deep base network appears to saturate.

\vspace{0.1cm}
\noindent \textbf{Size matching.}
Next, we study the influence of size matching methods on DCANet's performance. 
For matching the number of channels,
we evaluate the performance of the channel matching methods described in Section~\ref{section:size_matching}. 
We use SE-ResNet50 for illustration due to its pure concerns on channel dependencies. 
We utilize ``1 FC" to present direct matching and ``2 FC" to present two lightweight fully-connected layers.
Table~\ref{tab:ablation_size_matching_channel} presents the results. Directly applying one fully-connected layer can achieve the best top-1 accuracy, while, on the other hand, setting reduction rate $r$ to 16 in two fully-connected layers can reduce the number of parameters and achieve a comparable result. 
For spatial resolution, we adopt average pooling to reduce the resolution. We also compare with max pooling and present the results in Table~\ref{tab:ablation_size_matching_spatial}.  The performance of max pooling is slightly inferior compared to the performance of average pooling, indicating that all attention information should be passed to the succeeding attention blocks.

\vspace{0.1cm}
\noindent \textbf{Multiple Attention dimensions.}
Thus far, we comprehensively evaluate the performance of DCANet on a single attention dimension. Next, we describe the application on the attention modules, which take two or more attention dimensions into consideration, as presented in Section~\ref{section:multi_dimension}.  
For illustration, we use CBAM-ResNet50 as a baseline since CBAM module considers both channel-wise and spatial attentions. 
We first integrate DCANet with each attention dimension separately and then integrate them concurrently. 
We use DCA-C/DCA-S to present applying DCANet on channel/spatial attention, and DCA-All indicates we apply DCA module on both attention dimensions for CBAM-ResNet50.

Table~\ref{tab:ablation_multi-dimension} shows the results of DCANet applied on two dimensions. From the table, we notice that applying DCANet on either dimension will certainly improve the accuracy. From this, we note two interesting observations: 1) The improvements are slightly different: applying DCANet to channel-wise attention achieves 0.2\% more improvement than to spatial attention. 
2) Compared to the single dimension case, applying DCA to both attention dimensions achieves better performance. While the improvement of DCA-All is greater than either channel or spatial individually, it is less than the summation of the individual improvements as a whole. 
When enhancing both attention dimensions, we achieve a 0.54\% improvement. When we work on spatial and channel dimensions separately, the improvement is 0.51\% and 0.29\%, respectively.

\begin{figure}
\centering
\includegraphics[width=0.95\linewidth]{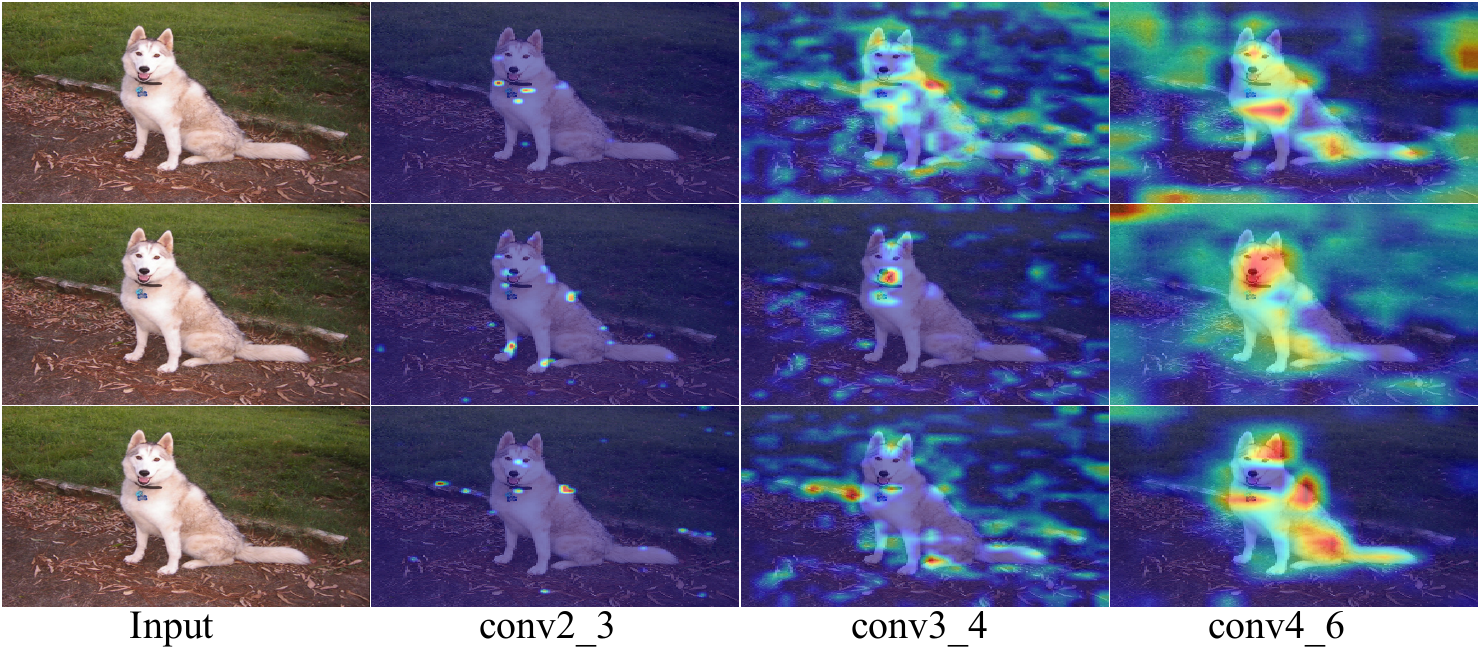}
\caption{Feature activation map visualization. \textbf{Top line:} ResNet50; \textbf{Middle line:} CBAM-ResNet50; \textbf{Bottom line:} DCA-CBAM-ResNet50. 
}
\label{fig:ablation_visualization}
\end{figure}


\subsection{Visualization}

In this section, we show that DCANet can adjust attention progressively for better feature extraction by visualization. 
We train CBAM-ResNet50 and DCA enhanced counterpart for 100 epochs on ImageNet. 
For DCA enhanced variant, we  connect attention modules along the spatial dimension to ensure the influence only comes from spatial dimension (referring DCA-S in Table~\ref{tab:ablation_multi-dimension}). We visualize intermediate feature activation using Grad-CAM~\cite{selvaraju2017grad}.  

Fig. \ref{fig:ablation_visualization} compares ResNet50, CBAM-ResNet50 and DCA counterpart.
\textbf{We see that the DCA-enhanced model consistently focuses on the key parts in an image, and at the same time, we also notice that the most discriminative parts vary little among intermediate features}. In contrast to our DCA enhanced model, feature maps from the vanilla CBAM-ResNet50 change drastically, as shown in the top row. Interestingly, the attention maps of CBAM-ResNet50 in Stage 3 are trivial and not discriminative while our DCA variant closely focuses on the key parts of the dog (as shown in the bottom row), indicating that DCANet  is able to enhance the learning power of attention module and further improve feature extraction power in CNN models.    





\subsection{Object detection on MS COCO}

We further evaluate the performance of DCANet for object detection. 
We measure the average precision of bounding box detection on the challenging COCO 2017 dataset~\cite{lin2014microsoft}. The input images are resized to make the shorter side to be 800 pixels~\cite{lin2017feature}.
We adopt the settings used in~\cite{mmdetection} and train all models with a total of 16 images per mini-batch (2 images per GPU).  We employ two state-of-the-art detectors: RetinaNet~\cite{lin2017focal} and Cascade R-CNN~\cite{cai2018cascade} as the detectors, with SE-ResNet50, GC-ResNet50 and their DCANet variants as the corresponding backbone respectively. All backbones are pre-trained using ImageNet and are directly taken from Table~\ref{table:comparison_ImageNet}. The detection models are trained for 24 epochs using synchronized SGD with a weight decay of 0.0001 and a momentum of 0.9. The learning rate is set to 0.02 for Cascade R-CNN and 0.01 for RetinaNet as previous work~\cite{SGENet,cai2018cascade,lin2017focal}. We reduce the learning rate by a factor of 10 at the 18$th$ and 22$nd$ epochs. 

\begin{table*}[!t]
\caption{Detection performances (\%) with different backbones on the MS-COCO validation dataset. We employ two state-of-the-art detectors: RetinaNet~\cite{lin2017focal} and Cascade R-CNN~\cite{cai2018cascade} in our detection experiments.}
\begin{center}
\begin{tabular*}{0.95\textwidth}{c @{\extracolsep{\fill}} cccccccc}
\toprule
Detector&Backbone  & AP$_{50:95}$ & AP$_{50}$ &AP$_{75}$ & AP$_S$ &AP$_M$ &AP$_L$  \\ 
\midrule
& ResNet50 &  36.2 &55.9 & 38.5 &19.4 & 39.8 & 48.3 \\
RetinaNet& + SE     &37.4 &57.8 &39.8  &20.6 &40.8 &50.3 \\
& + DCA-SE & \textbf{ 37.7}& \textbf{58.2}&   \textbf{40.1} &\textbf{20.8} & \textbf{40.9}& \textbf{50.4}\\ 
\midrule
& ResNet50     &40.6 &58.9 &44.2   &22.4 &43.7 &\textbf{54.7} \\
Cascade R-CNN& + GC &41.1 &59.7 &44.6   &\textbf{23.6} &44.1 &54.3 \\
& + DCA-GC &\textbf{41.4} &\textbf{60.2} &\textbf{44.7}   &22.8 &\textbf{45.0} &54.2 \\ 
\bottomrule
\end{tabular*}
\end{center}
\label{tab:detection}
\end{table*}

The results are reported in Table~\ref{tab:detection}. Although DCANet introduces almost no additional calculations, we observe that DCANet achieves the best performance for all IoU threshold values and most object scales (DCA-SE-ResNet50 obtains +1.5\% AP$_{50:95}$ on ResNet50 and +0.3\% AP$_{50:95}$ on SE-ResNet on SE-ResNet50 in RetinaNet; DCA-GC-ResNet50 obtains +0.8\% AP$_{50:95}$ on ResNet50 and 0.3\% AP$_{50:95}$ on GC-ResNet50 in Cascade R-CNN). 

Note that we only replace the backbone models by our methods, which means that all performance improvements come from our DCANet. The promising results indicate that our DCANet also works well for object detection tasks.

\subsection{Connection Analysis}
\begin{figure*}[!t]
\centering
\includegraphics[width=1\linewidth]{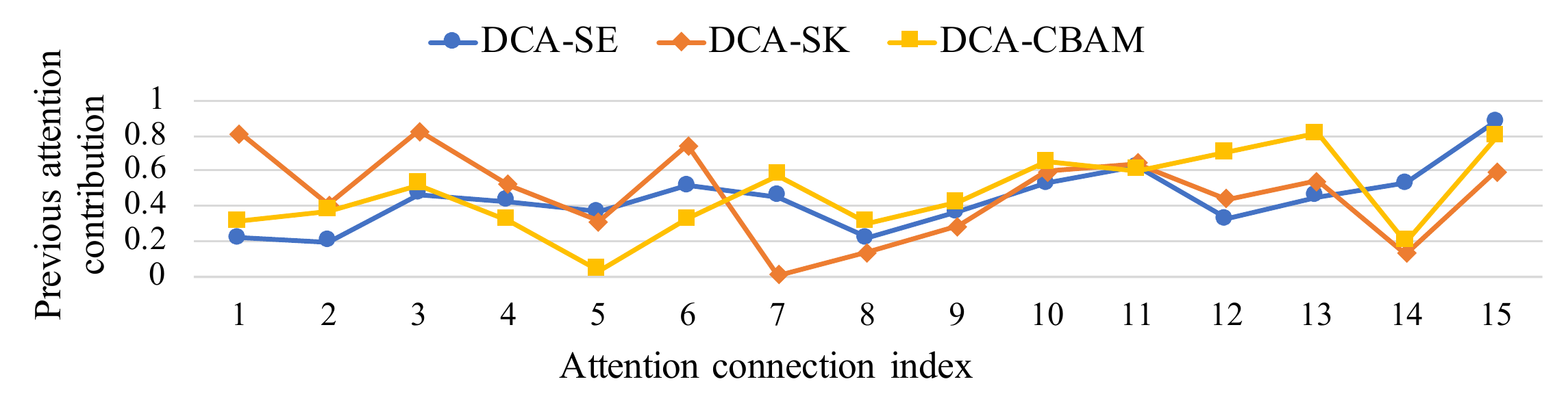}
\caption{The contribution ratio of the previous attention in each connection.}
\label{fig:weights}
\vspace{-0.1in}
\end{figure*}

To better understand how previous attention information contributes to attention learning, we measure the contribution ratio for each attention connection. We calculate the contribution ratio as $\beta/(\alpha+\beta)$. Thus, the contribution rate of previous attention falls  into a range of $\left[0,1\right]$. We take ResNet50 as a base network and  evaluate DCANet on SE, SK and CBAM modules. 

As shown in Fig.~\ref{fig:weights}, the contribution ratios are different from each other, showing that each attention module has its own intrinsic properties, and the contribution ratio does not follow a specific paradigm. Meanwhile, it is interesting to notice that from the 7$th$ to 13$th$ connections, the contribution ratios of all attention modules are stable compared to other connections. Markedly, the 7$th$ to 13$th$ connections are connections in The Stage 3 of a ResNet. The key insight behind this observation is that our DCANet would largely boost the feature extraction ability of later layers. 
Such an observation can also be confirmed in Fig.~\ref{fig:ablation_visualization}. in stark contrast to CBAM-ResNet50 and ResNet50, our DCA-CBAM-ResNet50 closely pays attention to the dog and has almost no activation on unrelated regions in stage 3. 
Second, the contribution ratios in the 14$th$ connection of all attention modules are close to 20\% and then enlarge in the last connection. Overall, the contribution ratio in all connections is always greater than 0, which means that the previous attention always contributes to the attention learning in the current attention block.

\section{Discussion}
DCA seems to  apply shortcut connections (as in ResNet) to attention blocks. Actually, they are fundamentally different. We connect attention blocks for joint training, whereas shortcuts in ResNet mitigate the ``degradation" problem. Besides, Fig.~\ref{fig:1} empirically reveals the difference. The shortcuts in SE-ResNet50 have almost no capability to facilitate the attention learning like our attention connections in DCANet.

More connection designs can be explored in DCANet, for example connecting each attention block to every other attention block in a dense fashion like DenseNet~\cite{huang2017densely}, or connecting attention blocks in a tree-like structure, similar to DLA~\cite{yu2018deep}. We will evaluate the performance of these connections in our future research.


We also note that connecting attention blocks deviates from self-attention which learns from the feature map itself. By connecting attention blocks, each attention block learns from both the previous attention map and the current feature map. The previous attention map can be considered as a guide for attention learning from the current feature map.

\section{Conclusion}
In this paper, we exhaustively point out that the capacity of self-attention mechanism is not fully explored. 
In order to explore better utilization, we present Deep Connection Attention Network (DCANet), which adaptively propagates the information flow among attention blocks via attention connections. 
We have demonstrated that DCANet consistently improves various attention designs and base CNN architectures on ImageNet benchmark with minimal computational overhead. Besides, experiments on MS-COCO datasets showcase that our DCANet generalizes well on other vision tasks. The elegant design and feed-forward manner of DCANet make it easy to be integrated with various attention designs using mainstream frameworks.



\bibliographystyle{splncs04}
\bibliography{egbib}
\end{document}


\pagestyle{headings}
\mainmatter
\def\ECCVSubNumber{4574}  

\title{DCANet: Learning Connected Attentions for Convolutional Neural Networks \\ Supplemental Materials} 

\titlerunning{ECCV-20 submission ID \ECCVSubNumber} 
\authorrunning{ECCV-20 submission ID \ECCVSubNumber} 
\author{Anonymous ECCV submission}
\institute{Paper ID \ECCVSubNumber}

\maketitle

This supplementary document mainly consists two parts. The first part presents the structure of DCANet and code link for reproducing our experiments. The second part provides visual
comparisons of more methods on more examples.

\section{Structure \& Code}
Our Deep Connected Attention Network (DCANet) is general in nature and can be used on various attention modules and base networks. Without DCANet, there is only one connection between two adjacent blocks and the next block learns from the output of previous block. By incorporating DCANet, convolutional block not only learns from the output of its previous block, but also takes previous attention features into consideration. This further highlights the reuse of attention features in an attention block which contributes for feature extraction. Besides, the implicitly learning from residual connections~\cite{he2016deep} can barely help the attention learning. 

The source code of our DCANet is publicly available at GitHub. The link is \href{https://github.com/eccv2020-4574/DCANet}{https://github.com/eccv2020-4574/DCANet}. We also provide the trained models and training log files in the repository. We welcome feedback and suggestions, and hope that by releasing the code and models publicly, we can provide a better interpretation of our Deep Connected Attention Networks.

\section{Visualization Results}
We plot more attention maps to visualize the effects of DCANet. We employ several attention modules, including SKNet\cite{li2019selective}, CBAM\cite{woo2018cbam}, and SENet\cite{hu2018squeeze}. Fig. \ref{fig:sup_attention} shows the visualization results. The results indicate that regardless of the choice of attention module and base network, our DCANet always can progressively adjust network focus.

We also present the detection results of DCANet for object detection task. Fig. \ref{fig:detection_results} compares the detection results of our DCANet enhanced backbone and vanilla backbone.  Across all 4 columns of detection results we observe that our DCANet enhanced networks are more likely to achieve a compact bounding box, and further achieve a better detection performance.

\begin{figure*}
    \centering
    \includegraphics[width=1\linewidth,height = 14.5cm]{images/sup_v.pdf}
    \caption{Visualization of attention maps. From top to bottom are results generated by SK-ResNet50\cite{li2019selective,he2016deep}, DCA-SK-ResNet50, SE-MnasNet1\_0\cite{hu2018squeeze,tan2019mnasnet}, DCA-SE-MnasNet1\_0, CBAM-MobileNetV2\cite{woo2018cbam,sandler2018mobilenetv2} and DCA-CBAM-MobileNetV2. }
    \label{fig:sup_attention}
\end{figure*}

\begin{figure*}
    \centering
    \includegraphics[width=0.23\linewidth, height=3.3cm]{images/det/retina_se/det_0.jpg}
    \includegraphics[width=0.23\linewidth, height=3.3cm]{images/det/retina_dca/det_0.jpg}
    \includegraphics[width=0.23\linewidth, height=3.3cm]{images/det/cascade_gc/det_0.jpg}
    \includegraphics[width=0.23\linewidth, height=3.3cm]{images/det/cascade_dca/det_0.jpg}

    \includegraphics[width=0.23\linewidth, height=3.3cm]{images/det/retina_se/det_3.jpg}
    \includegraphics[width=0.23\linewidth, height=3.3cm]{images/det/retina_dca/det_3.jpg}
    \includegraphics[width=0.23\linewidth, height=3.3cm]{images/det/cascade_gc/det_3.jpg}
    \includegraphics[width=0.23\linewidth, height=3.3cm]{images/det/cascade_dca/det_3.jpg}

    \includegraphics[width=0.23\linewidth, height=3.3cm]{images/det/retina_se/det_6.jpg}
    \includegraphics[width=0.23\linewidth, height=3.3cm]{images/det/retina_dca/det_6.jpg}
    \includegraphics[width=0.23\linewidth, height=3.3cm]{images/det/cascade_gc/det_6.jpg}
    \includegraphics[width=0.23\linewidth, height=3.3cm]{images/det/cascade_dca/det_6.jpg}

    \includegraphics[width=0.23\linewidth, height=3.3cm]{images/det/retina_se/det_9.jpg}
    \includegraphics[width=0.23\linewidth, height=3.3cm]{images/det/retina_dca/det_9.jpg}
    \includegraphics[width=0.23\linewidth, height=3.3cm]{images/det/cascade_gc/det_9.jpg}
    \includegraphics[width=0.23\linewidth, height=3.3cm]{images/det/cascade_dca/det_9.jpg}
    
    \includegraphics[width=0.23\linewidth, height=3.3cm]{images/det/retina_se/det_1.jpg}
    \includegraphics[width=0.23\linewidth, height=3.3cm]{images/det/retina_dca/det_1.jpg}
    \includegraphics[width=0.23\linewidth, height=3.3cm]{images/det/cascade_gc/det_1.jpg}
    \includegraphics[width=0.23\linewidth, height=3.3cm]{images/det/cascade_dca/det_1.jpg}
    
    \caption{Detection results on MS COCO dataset \cite{lin2014microsoft}. From left to right are detection results of RetinaNet (backbone: SE-ResNet50), RetinaNet (backbone: DCA-SE-ResNet50), Cascade R-CNN (backbone: GC-ResNet50), and Cascade R-CNN (backbone: DCA-GC-ResNet50).}
    \label{fig:detection_results}
\end{figure*}

\bibliographystyle{splncs04}
\bibliography{egbib}